# OSTEOPOROTIC AND NEOPLASTIC COMPRESSION FRACTURE CLASSIFICATION ON LONGITUDINAL CT

*Yinong Wang[1], Jianhua Yao[1], Joseph E. Burns[2], Ronald M. Summers[1]*

[1]Imaging Biomarkers and Computer-Aided Diagnosis Laboratory, Radiology and Imaging Sciences, Clinical Center, National Institutes of Health
[2]Department of Radiological Sciences, University of California-Irvine, School of Medicine

## ABSTRACT

Classification of vertebral compression fractures (VCF) having osteoporotic or neoplastic origin is fundamental to the planning of treatment. We developed a fracture classification system by acquiring quantitative morphologic and bone density determinants of fracture progression through the use of automated measurements from longitudinal studies. A total of 250 CT studies were acquired for the task, each having previously identified VCFs with osteoporosis or neoplasm. Thirty-six features for each identified VCF were computed and classified using a committee of support vector machines. Ten-fold cross validation on 695 identified fractured vertebrae showed classification accuracies of 0.812, 0.665, and 0.820 for the measured, longitudinal, and combined feature sets respectively.

## 1. INTRODUCTION

Compression fractures of the vertebral body (VCF) are highly prevalent in individuals over the age of 50, with a predisposition for females due to their inherently lower bone density compared to their male counterparts [1]. Such occurrences manifest as benign or malignant fractures that result from osteoporotic and neoplastic origins, respectively (Figure 1) [2]. VCFs can produce substantial pain and movement difficulty, and may follow a course of further compression. Diagnosis of VCFs is typically evaluated through qualitative visual review of height loss and bone density through imaging modalities such as radiography and computed tomography (CT). Identifying the etiology of VCF development is fundamental to treatment planning due to the markedly different methodologies used to treat neoplastic and osteoporotic VCFs, ranging from conservative management such as bracing to more invasive measures such as fixation hardware or radioactive cement placement.

Factors leading to the development of vertebral compression fractures have been extensively evaluated in the clinical setting. Morphological parameters of vertebrae including the vertebral body height have been examined using post-mortem examinations and physical measurements of normal and healthy adult vertebral column specimens [3]. Normative databases have been developed for measurements of vertebral height and other parameters from manually designated computer-aided measurements on radiographic views of the spine [4]. In addition to changes in vertebral body height, correlation between the trabecular bone density and compression strength suggests that the measurement of the bone density via imaging modalities may provide insight towards estimating the likelihood of compression [5]. Vertebral compression fractures have also been shown to be a substantially important predictive factor for subsequent fracture risk due to the compounding nature of biomechanical failure of the spine [6].

Despite the extensive amount of interest in identifying factors that contribute to vertebral compression fractures, existing clinical decision-making paradigms for the planning of VCF treatment have been hindered by a lack of quantitative morphologic and bone density determinants of fracture progression. By monitoring changes in vertebra height and bone mineral density, we measure differences that may exist between vertebrae with osteoporotic and neoplastic compression fractures on CT. Using existing computational techniques for measuring bone density and local and global descriptors of vertebral body height, we outline the construction of a model for classifying osteoporotic and neoplastic origin expressed by identified fractured vertebrae.

## 2. METHODS

The framework for the classification of osteoporotic and neoplastic vertebral compression fractures was accomplished by measuring features of vertebral body height and bone density on CT over the span of multiple studies per patient. An automatic method was used on CT to segment the spinal column and partition each individual vertebra and allowed global and local descriptors of height and measures of cortical and trabecular bone density to be obtained. The rate of change for each measured feature, denoted as longitudinal features, were determined using the time elapsed between studies. These values were then passed to a committee of support vector machines (SVM) for the classification task.

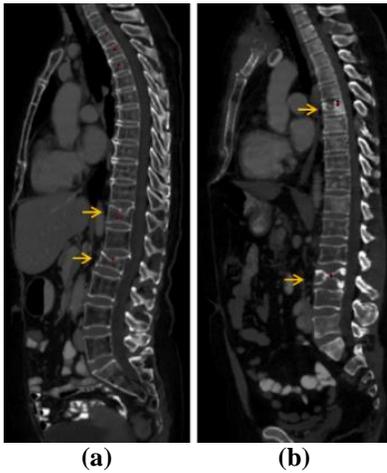

**Figure 1.** Sagittal view of vertebral compression fractures (arrows) of **(a)** osteoporotic and **(b)** neoplastic origin on CT.

### 2.1. Spine Segmentation

The extraction of features used for classification first requires segmentation of the spine. This was achieved by using an automated method for segmenting the spinal column and partitioning the vertebrae (Figure 2) [7]. The spinal canal was segmented using adaptive thresholding, watershed, and directed graph search. An anatomic vertebra model and curved reformations were used to identify and partition individual vertebrae.

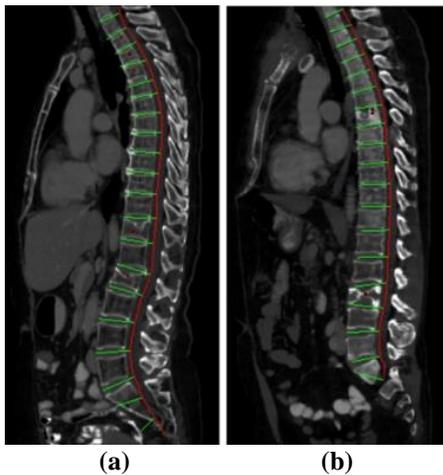

**Figure 2.** Sagittal view of partitioned thoracolumbar vertebrae in patients with **(a)** osteoporosis and **(b)** neoplasm on CT.

### 2.2. Height Measurement

A group of features pertaining to the height of the vertebral body was computed by using a height compass (Figure 3) [8]. The compass partitioned each vertebral body axially into 17 cells oriented in concentric rings with eight equal length arcs (Figure 3). The superior and inferior endplates of each vertebra were identified and the distance between them was computed in all 17 cells. Features of height (mm) were summarized as mean measurements across the central ($h\_c$), axial ($h\_a$), posterior ($h\_p$), left ($h\_l$), and right ($h\_r$) regions of the vertebral body, as well as an overall mean ($h\_avg$). The level of each vertebra (vid) and the relative height of the vertebra of interest with respect to its adjacent vertebrae (contrastP, contrastN, and contrastA) were also recorded. The heights of the center, anterior and posterior edges, and the mean height of the vertebral body were also measured in a mid-line sagittal view.

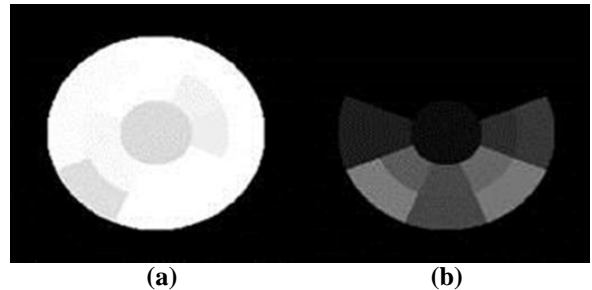

**Figure 3.** Height compass layout and orientation in the axial plane of a **(a)** normal and **(b)** fractured vertebral body.

### 2.3. Bone Density Estimation

The bone density was estimated using automated placement of a region-of-interest generated from the intensity-based segmentation of each vertebra [9]. A mean Hounsfield number (HU) was calculated using the segmentation. By eroding the segmentation of the vertebral body to the anterior half as a method to remove the cortical bone, a mean Hounsfield number was determined for the trabecular bone using the remaining volume. The estimations were normalized using segmentations of the muscle and fat near the spinous process (Figure 4). Bone density was summarized as the mean cortical density (meanDen) and mean trabecular density (meanTrab).

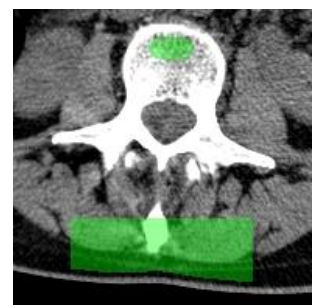

**Figure 4.** Bone density estimation in the anterior region of the vertebral body. Segmentations of the muscle and fat near the spinous process were used for the normalization of the determined bone density.

## 2.4. Longitudinal Features

The use of height and density features from successive studies introduced the ability to examine the change in those features over time (Figure 5). The rate of change in all height (mm/year) and density (HU/year) features were computed using the measured value from the current and previous time points normalized over the period of time elapsed between studies.

## 2.5. Feature Selection and Classification

A total of thirty-six features were collected and forwarded to a feature selection program to determine the best groups of features for classifying osteoporotic and neoplastic vertebral compression fractures using a committee of support vector machines, shown in Table 1 [10, 11]. We grouped the features into measured features (height and density features measured for each study) and longitudinal features (rate of change in height and density features computed between the current and previous study). Two patient demographic features (gender and age) were also included as part of the measured and longitudinal feature sets. Training data for the classification of a compression fracture as osteoporotic or neoplastic was generated by previous visual inspection of each vertebra and study by a board-certified radiologist. Performance was evaluated by ten-fold cross-validation.

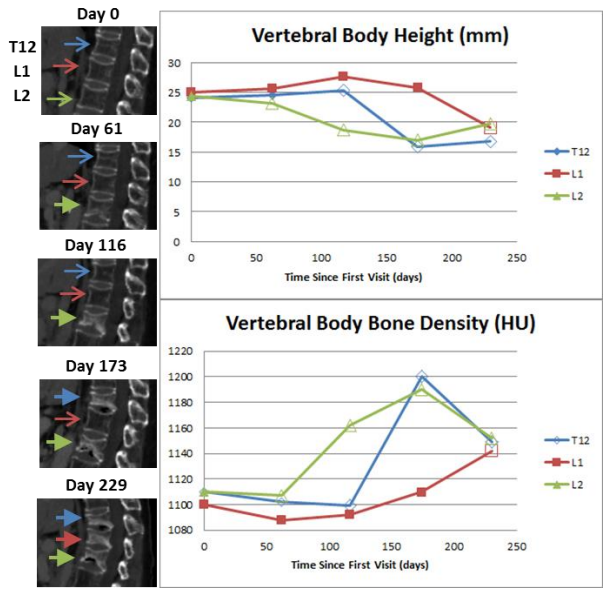

**Figure 5.** Longitudinal change of vertebral body height (mm) and bone density (HU) over the course of multiple studies on CT.

## 3. DATASET

A total of 250 CT studies containing scans of vertebral compression fractures (174 osteoporotic and 76 neoplastic) from a cohort of 56 patients (mean age 57±15 years, 27 female and 29 male) were used to build the classification model. Studies were retrospectively retrieved from the local PACS system and filtered based on a text search for "vertebra" and "compression fracture" with the additional requirement of having chest/abdomen/pelvis scans of slice thickness smaller than 2mm. Fractured vertebrae were identified upon visual review by a trained technician. Patients had an average of 6.2 studies acquired over a period of 1 month to 3.8 years. In total 695 vertebrae were identified with fractures (490 osteoporotic and 205 neoplastic).

**Table 1.** Summary of features for classification.

|  | Height | Bone Density |
|---|---|---|
| **Measured Features** | $h\_c$, $h\_a$, $h\_p$, $h\_l$, $h\_r$, $h\_{avg}$, $h\_{avg\_5}$, contrastP, contrastN, contrastA, vid, Anterior, Center, Posterior, manualMean, meanH | meanDen, meanTrab |
| **Longitudinal Features** | $R_{h\_c}$, $R_{h\_a}$, $R_{h\_p}$, $R_{h\_l}$, $R_{h\_r}$, $R_{h\_{avg}}$, $R_{h\_{avg\_5}}$, $R_{contrastP}$, $R_{contrastN}$, $R_{contrastA}$, $R_{Anterior}$, $R_{Center}$, $R_{Posterior}$, $R_{manualMean}$ | $R_{meanDen}$, $R_{meanTrab}$ |
| **Demographic Features** | Gender, Age | |

## 4. RESULTS

The classification accuracy for the measured feature set, longitudinal feature set, and combined feature set (containing all features) are 0.812, 0.665, and 0.820, respectively. Table 2 lists the associated confusion matrices. The longitudinal feature set produces a significantly higher number of misclassifications, 233 compared to 131 and 125 for the measured and combined feature sets respectively. Fisher's exact test showed that the performance of both the measured feature set and the combined feature set were statistically improved ($p<10^{-3}$) over the longitudinal feature set, but not between each other (p=0.665). Further analysis of the three sets of features shows that all methods underestimate the number of osteoporotic fractures and thereby overestimate the number of neoplastic fractures.

Examples of correct osteoporosis and neoplasm classification are shown in Figures 7a and 7d respectively. Misclassification of osteoporosis as neoplasm in Figure 7b is likely the result of the injection of medical cement to prevent further vertebral compression. Region-of-interest placement on the anterior half of the vertebral body is a probable contributor to the misclassification of Figure 7c as osteoporosis. The sites of metastatic disease and elevated bone density are located in the posterior region of the vertebral body, and are not captured by the algorithm.

## 5. CONCLUSION

We present a technique for the acquisition of features for the classification of vertebral compression fractures of osteoporotic and neoplastic origin. The data shows that the longitudinal feature set produces significantly more misclassifications than the other feature sets. However, the inclusion of longitudinal features for our classification using a committee of support vector machines may provide some benefit to classification accuracy, but improvements are not statistically significant.

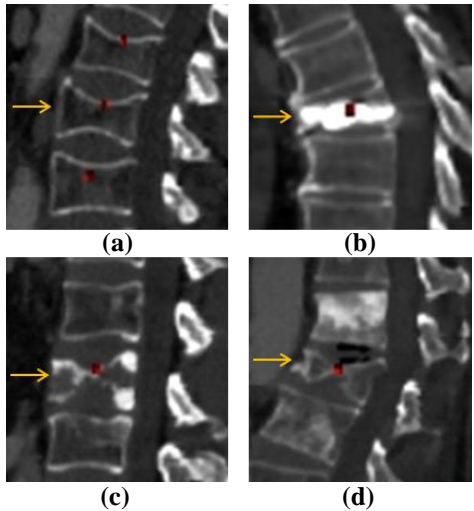

**Figure 7.** Classification results on vertebra of interest (arrow): **(a)** successful classification as osteoporotic, **(b)** misclassification as neoplastic, **(c)** misclassification as osteoporotic, and **(d)** successful classification as neoplastic.

**Table 2.** Confusion matrices for measured features, longitudinal features, and combined features. Radiologist diagnosis (row) and SVM classification (column) displayed.

| Measured | O | N | Total |
|---|---|---|---|
| O | 392 | 98 | 490 |
| N | 33 | 172 | 205 |
| Total | 425 | 270 | 695 |

| Longitudinal | O | N | Total |
|---|---|---|---|
| O | 345 | 145 | 490 |
| N | 88 | 117 | 205 |
| Total | 433 | 262 | 695 |

| Combined | O | N | Total |
|---|---|---|---|
| O | 399 | 91 | 490 |
| N | 34 | 171 | 205 |
| Total | 433 | 262 | 695 |

## 6. ACKNOWLEDGMENTS

This research was supported in part by the Intramural Research Program of the National Institutes of Health, Clinical Center.